\documentclass[letterpaper, 10 pt, conference]{ieeeconf}  
\IEEEoverridecommandlockouts                              
\usepackage{graphics} 
\usepackage{epsfig} 
\usepackage{amsmath} 
\usepackage{amssymb} 
\usepackage{hyperref}
\usepackage{algorithm}
\usepackage{algorithmic}
\DeclareMathOperator*{\argmax}{arg\,max}
\DeclareMathOperator*{\argmin}{arg\,min}

\usepackage{graphicx}
\usepackage{caption,subcaption}
\title{\LARGE \bf
	A Scavenger Hunt for Service Robots
}

\author{Harel Yedidsion$^{1}$, Jennifer Suriadinata$^{1}$, Zifan Xu$^{2}$, Stefan Debruyn$^{1}$, and 
	Peter Stone$^{1,3}$ 
	\thanks{$^{1}$
		The University of Texas at Austin, Department of Computer Science
		$^{2}$
		The University of Texas at Austin, Department of Physics
		$^{3}$
		Sony AI
		{\tt\small harel, pstone@cs.utexas.edu, jsuriadinata, zfxu, stefandebruyn@utexas.edu}%
	}
}

\begin{document}

	\maketitle
	\thispagestyle{empty}
	\pagestyle{empty}

	\begin{abstract}
		Creating robots that can perform general-purpose service tasks in a human-populated environment has been a longstanding grand challenge for AI and Robotics research. 
		One particularly valuable skill that is relevant to a wide variety of tasks is the ability to locate and retrieve objects upon request. 
		This paper models this skill as a Scavenger Hunt (SH) game, which we formulate as a variation of the NP-hard stochastic traveling purchaser problem.  In this problem, the goal is to find a set of objects as quickly as possible, given probability distributions of where they may be found.  
		We investigate the performance of several solution algorithms for the SH problem, both in simulation and on a real mobile robot. We use Reinforcement Learning (RL) to train an agent to plan a minimal cost path, and show that the RL agent can outperform a range of heuristic algorithms, achieving near optimal performance.
		In order to stimulate research on this problem, we introduce a publicly available software stack and associated website that enable users to upload scavenger hunts which robots can download, perform, and learn from to continually improve their performance on future hunts.

	\end{abstract}

	\section{INTRODUCTION}
	
	The scavenger hunt problem is concerned with optimizing the search for objects whose positions are uncertain. Given a list of objects to find alongside a probability distribution model describing where objects tend to be located, an autonomous agent tries to minimize the distance traveled to find all objects in the hunt. Objects may be located differently in different hunts, and these differences are unknown to the hunting agent. 
	
	This problem is difficult for a number of reasons, including the fact that it involves probabilistic events. It is more complex than the traveling salesman problem~\cite{reinelt2003traveling} because the potential rewards for visiting locations change as a scavenger hunt progresses and objects are found. Additionally, the shortest path through unvisited locations does not necessarily minimize the expected cost of hunt completion.
	
	A solution to the scavenger hunt problem has many applications for domestic service robots. Robots which can quickly search for objects could remove tedium from day-to-day activities. 
	Examples of real life scavenger hunts include grocery shopping, packing a suitcase, preparing a meal, or delivering lunch boxes to workers in an office. 
	
	In this paper we introduce and evaluate several algorithms for solving the scavenger hunt problem. The exhaustive Bayesian search algorithm uses Bayesian search theory~\cite{stone1976theory} to identify the search path with the lowest expected cost among all possible paths. It is however intractable for large instances, and not applicable for service robots that need to react quickly. 
	As a consequence, we devise three heuristic algorithms: a proximity-based search, a probability-based search, and a combined probability-proximity-based search algorithm.
	
	In addition, we extend this search under probabilistic events to the context of Reinforcement Learning (RL). We show that the problem can be modeled within the framework of partially observable Markov decision process (POMDP)~\cite{kaelbling1998planning}, which treats the true locations of the objects as latent states. Lizotte et al.~\cite{lizotte2008missing} solved a similar problem with missing data as latent states by maintaining a belief state over the posterior distribution of latent states. Then a Q-learning algorithm was used to learn a policy conditioned on the belief state instead of an incomplete observation. Inspired by this work, we represent a scavenger hunt as a fully observable MDP which always maintains the state as a posterior distribution over object locations. Then we solve the MDP with the standard Deep Q-network (DQN) algorithm~\cite{mnih2015human}.  We demonstrate that a learning-based algorithm can learn near-optimal solutions without requiring exponential runtime when deployed.	
	
	These algorithms, as well as two bounding algorithms, were implemented and tested in an abstract simulation of the scavenger hunt problem, and their performances over many randomized scavenger hunts were compared. Three of our algorithms were also implemented on a physical robot.
	The robot performed object location tasks around our laboratory space using autonomous navigation, motion planning, computer vision software for object recognition~\cite{redmon2016you}, and communication with our newly created scavenger hunt webserver, as part of real life scavenger hunts.

	One of our main goals is to present the Scavenger Hunt (SH) as a testbed for robot development. To this end, we developed a publicly available website that allows users around the world to compose hunts and validate their successful completion. We also developed a publicly available software stack that enables robots to download scavenger hunts and upload proofs of completion to the website.
	\footnote{The website url is:  \url{http://scavenger-hunt.cs.utexas.edu/} It contains a demo of our scavenger hunt system, the code for running the robot, and the simulations.}
	

	The contributions of this paper include: (i) the formulation of the robot scavenger hunt problem, (ii) the development and evaluation of optimal, heuristic, and learning-based algorithms to solve the scavenger hunt problem, (iii) the development of a publicly available website that forms the infrastructure for running scavenger hunts on robots and for providing feedback on their performance, and (iv) a demonstration of the complete system on a real mobile autonomous robot.
	
	
	\section{Background and Related Work}
	\label{sec:bg}
	
	Scavenger hunt is a challenging and fun way of testing robot skills which are essential for long term autonomy such as planning, navigation, continual lifelong learning, perception, and human robot interaction. In AI conferences, scavenger hunts have been used as a benchmark for autonomous robot performance in public events where robots were required to search a cluttered environment for a defined list of objects~\cite{casey2005good,rybski2006aaai, rybski2007aaai}. 
	
	The scavenger hunt task has also been proposed as a standardized benchmark for autonomous robots by several research labs~\cite{IJCAI16-szhang}. 
	We focus on a version of scavenger hunt which is most similar to the target search task described by Zhang et, al.~\cite{IJCAI16-szhang}, and present a website to coordinate cooperation between academic labs.
	
	We define the robot scavenger hunt problem as a variant of the well-studied traveling purchaser problem (TPP)~\cite{manerba2017traveling}.
	The TPP is defined as follows. Consider a set $O$ of products/items to purchase, and a set $N$ of geographically dispersed suppliers to choose from. A solution is a path visiting a subset of suppliers, and a decision of which items to purchase from each supplier so as to minimize the traveling and purchasing costs.
	The specific TPP variant we consider in this paper, is in some sense more complex than the classic TPP since the product availability is non-deterministic, and only a distribution model is available. On the other hand, it is simpler since we do not consider purchasing quantities or costs. 
	The SH problem is similar to the Multi-Object Search problem considered by Wandzel et al. \cite{wandzel2019multi}, except that their proposed planning-based method does not generalize to arbitrary object distributions.

	In the general case of solving a NP-hard combinatorial optimization problem, such as TSP, RL has been successfully employed to learn a policy that achieved near-optimal performance and generalized well to different graphs and sizes~\cite{NIPS2017_d9896106}. However, our problem considers a more complex setting with stochasticity in object locations. 
	In a similar problem known as continual area sweeping ~\cite{ahmadi,shah2020deep}, 
	an RL approach allowed the robot to detect events and learn their distributions simultaneously. 
	The scavenger hunt problem differs in that the agent has to plan over a known probability distribution of object locations.
	Previous work using a similar method explores problems with a limited amount of missing data or data with small Gaussian noise ~\cite{lizotte2008missing,DBLP:journals/corr/abs-1902-05795}, while the SH problem requires considering all the object locations under uncertainty, and the probability distribution is arbitrary. 	
	
	In the work by Lau et al.~\cite{lau}, a simulated agent searches for objects whose locations vary according to a probability distribution. The dynamic programming approach used by their simulated agent is very similar to our Bayesian search algorithm. The objects, however, had no distinct identities, and their locations varied according to the same distribution, unlike the SH problem, where each object has a unique identity and probability distribution.
	
	Other projects proposed efficient methods for modeling and updating the distribution of moving object locations \cite{sung2017algorithm} while our model considers stationary object locations which are randomly sampled at the beginning of the search process.
	
	\section{Problem Formulation}
	\label{sec:problem_formulation}
	
	We denote a connected graph by $G = \{N, E\}$, where $N = \{n_1, n_2, \dots, n_l\}$ is the node set and $E = \{e_{n, n'}\}_{n, n' \in N}$ is the cost set with any $e_{n, n'} \in E$ representing the minimal cost of traveling from node $n$ to node $n'$.
	A scavenger hunt problem consists of a graph $G = \{N, E\}$, an object set $O = \{o_1, \dots, o_k\}$, a prior distribution $D$ over $N^{k}$, and a starting node $n_0 \in N$.
	
	A scavenger hunt problem first generates a vector of object locations $X = (X_{o_1}, \dots X_{o_k}) \in N^{m}$ from the prior distribution $D$, where $X_{o_{i}}$ represents the location of object $i$ for $i \in 1, \dots, k$. The object locations $X$ are unobservable to the searching agent. We denote $t$ as the timestep and let $Y_t \in \{0, 1\}^k$ be the task vector at timestep $t$ , where $Y_{t,i} = 0$ if $o_i$ has not been found and $Y_{t,i} = 1$ otherwise.
	An algorithm for solving a scavenger hunt problem starts from node $n_0$ and initial task vector $Y_0 = (0, 0, \dots, 0)$. At each timestep $t \geq 1$, the searching agent takes an action by moving to node $n_t$, and incurring a cost $e_{n_{t-1}, n_{t}}$.
	
	An agent's performance is evaluated by the total cost before all $Y$'s are set to $1$. Formally, the objective of a SH problem is to minimize the cost as defined by Equation \ref{eq:objective}:
	
	\vspace{-10pt}
	\begin{equation}
		\label{eq:objective}
		\begin{aligned}
			\min \quad & \sum_{t = 1}^{t_1} e_{n_{t-1}, n_t}, \text{ for } t_1 = \inf \{t > 0: Y_{t, i} = 1, \forall i \in [k]\}\\
			\textrm{given} \quad & G,D,O,n_0\\
		\end{aligned}
	\end{equation}


	
	Algorithm \ref{alg:framework} describes a general framework for all the heuristic scavenger hunt algorithms in section 4. 
	The algorithms differ in the their implementation of the $choose\_next\_node()$ function on line \ref{line:choose}. On line \ref{line:update}, the agent updates the task vector $Y_t$ and a trajectory $\tau_t = (n_0, Y_0, e_{n_0, n_1}, n_1, Y_1, e_{n_1, n_2}, \dots, n_{t-1}, Y_{t-1})$ up to time $t$ after each timestep. A Bayesian posterior distribution $D(X|\tau_t)$ based on the prior distribution $D=D(X|\tau_0 = \emptyset)$ and the trajectory $\tau_t$ is computed for the algorithm to make a decision. 
	
	\begin{algorithm} 
		
		\caption{A General Framework for SH Algorithms}
		\label{alg:framework}
		\begin{algorithmic}[1]
			\REQUIRE{A graph $G = (N,E)$, a set of Objects $O$, a distribution $D$, start location node $n_0$}
			\STATE{Initialize $n_0$, $Y_0$ and $\tau_0 = \emptyset$ with $t = 0$}
			\WHILE{$Y_{t, i} \neq 0, \forall i \in [k]$}
			\STATE{$n_{t+1}$ $\xleftarrow{}$ choose\_next\_node(G,D,O,$n_t$,$\tau_t$)} \label{line:choose}
			\STATE{Travel to $n_{t+1}$}
			\STATE{Update $D$, $Y_{t+1}$ and $\tau_{t+1}$ based on the occurrence of objects at $n_{t+1}$}
			\label{line:update}
			\STATE{$t = t+1$}
			\ENDWHILE  
		\end{algorithmic}
		
	\end{algorithm}
	
	\section{Solution Algorithms}
	\label{sec:algorithms}
	In this section we present seven scavenger hunt-solving algorithms: one RL algorithm (DQN), one exhaustive search algorithm (Bayesian), three heuristic algorithms (Proximity, Probability, and Probability-Proximity), and two bounding algorithms (Offline Optimal, and Salesman). 
	Complete pseudocode for all the algorithms is available at: {\small \url{https://github.com/utexas-bwi/scavenger_hunt_api/blob/master/PseudoCode.pdf}}
	
	
	\subsection{DQN Algorithm (RL)}
	
	While the system is originally partially observable to the agent due to the unknown object locations, maintaining a dynamically updated Bayesian posterior distribution of the object positions as the state, leads to a regular MDP. 
	Therefore, the algorithm considers the SH problem as an MDP represented as a tuple $(S, A, P, r, \gamma)$, with state-space $s\in S$: a state $s$ is a set \{$D$, $n$\}, which includes the probability distribution $D$, combined with agent's current node $n$. 
	The probability distribution is represented by a $l\times k$ matrix with element $p_{t_o}^n$ representing the probability of object $o$ being located at node $n$; action-space $a \in A$: $A$ is the same as the node set $N$ with an action $a = n$ representing an action of travelling to node $n$; transition probability of states $P := p(s_{t+1}|s_t, a_t)$:  a probability of transitioning from state $s_t$ to the next state $s_{t+1}$ given the action $a_t$;
	reward function $r:=r(s,a)$: the negative distance of travelling from the agent's current node to the node specified by the action; a discount factor $\gamma = 0.95$ is set only for the purpose of training. 
	
	DQN with experience replay is implemented to train an agent with respect to just one specific hunt instance at a time. The Q-network is represented as a multi-layer perceptron with two 16-unit layers. Each of the layers is followed by a rectified linear unit (ReLu) activation function. The network takes as input a vector of probabilities of finding at least one object at each node, together with the shortest path costs between nodes.
	As we defined before, $p_t^{n}$ is the probability of finding at least one object at node $n$ and time step $t$ with $n \in \{n_j\}$ for index $j = 1, 2,\dots, l$ and we denote $n_t$ as agent's current node, a vector $(p^{n_1}_t, p^{n_2}_t, \dots , p^{n_l}_t, e_{{n_1}, n_t}, e_{{n_2}, n_t}, \dots, e_{{n_l}, n_t})$ of length $2l$ is sent to the network as input. The agent's current node is apparent as being the only node with 0 travel cost.
	The network outputs the expected Q-values for each of the actions. The policy $\pi(a|s):=\argmax_{a \in A} Q(s,a)$ is extracted by selecting the action that returns the highest Q value. 
	
	The hyper-parameters used for the training are specified as follows. The weights of the network are optimized by Adam optimizer~\cite{kingma2014adam} a linearly decaying learning rate starting from 0.05 and reaching 0 after 40 epochs. The agent is trained for 2000 steps per epoch with a batch size of 64 and is tested for 200 episodes after each epoch. A buffer size of 20000 is used for the experience replay and a epsilon-greedy policy is introduced to do exploration during the training with epsilon linearly decaying at a rate of 0.1 per epoch and staying at 0.02 after 10 epochs. Decaying the learning rate and epsilon help encourage convergence to a near optimal policy. The policy with the best test result is reported as the result of the algorithm. 
	
	\subsection{Exhaustive Bayesian Search (Exhaustive)}
	
	
	This algorithm considers all possible paths, which are the sequences that contains all the unvisited nodes, and computes the expected cost of each path over all possible object locations $X \in N^k$.  The path with the lowest expected cost over the distribution $D$ is selected and the first node in the path is returned as the next node to visit. 
	
	
	
	

	\subsection{Proximity-Based Search (Heuristic)}
	
	This algorithm chooses the next node to visit by searching for the closest node that may potentially contain any unfound object. The number of objects that may appear at a node and probability of appearance are disregarded. We denote ${p_t}_o^{n} := D(X_o = n|\tau_t)$ as the posterior probability of object $o$ located at node $n$ based on the trajectory $\tau_t$. 
	The posterior probability incorporates past observations, including whether an object has already been found.
	Given the robot's current node $n_t$, the next node $n_{t+1}$ is computed by the following equation: 
	\begin{equation}
		n_{t+1} = \argmin\limits_{n \in N} \{e_{n_t, n}|\exists~o \in O, {p_t}_o^{n} > 0\}
	\end{equation}

	%
	

	\subsection{Probability-Based Search (Heuristic)}
	
	This algorithm chooses the node with the greatest probability of finding at least one unfound object. The proximity of the location is disregarded, resulting in essentially the opposite of the proximity-based heuristic. If we denote $p_t^n$ as the probability of finding at least one object at node $n$, the next node is computed by: 
	
	\begin{equation}
		\begin{split}
			n_{t+1} = \argmax\limits_{n\in N} \{p_{t}^{n}\} \quad\text{and}\quad 
			p_{t}^{n} = 1-\prod\limits_{o \in O} (1-{p_t}_o^n)
		\end{split}
	\end{equation}

	
	\subsection{Probability-Proximity Search (Heuristic)}
	This algorithm is a combination of the proximity-based and probability-based heuristics. Each location node is scored according to the ratio of the probability of finding at least one object at that node to the distance from the current node to that node.
	Given the current node $n_t$ and the probability of finding at least one object $p_t^n$, the next node is computed by the following equation:
	\begin{equation}
		n_{t+1} = \argmax\limits_{n\in N} \frac{p_t^{n}}{e_{n_t, n}}
	\end{equation}
	
	The algorithm visits the node with the highest ratio. The rationale behind this scaling is to prioritize visiting nodes if the cost of traveling to them is worth the expected reward.

	
	\subsection{Salesman Search (Bounding)}
	Serving as a naive baseline indicator on performance is a traveling salesman algorithm which visits all nodes that may contain objects, along the shortest path, while disregarding occupancy distributions~\cite{reinelt2003traveling}. The path is calculated once and not reevaluated during the search.
	This algorithm is intended as a lower bound on the performance of other heuristics, however, it is not a strict lower bound since for specific locations of objects, it can outperform other heuristics. 
	
	\subsection{Offline Optimal Search (Bounding)}
	
	The offline optimal search is informed offline of the objects locations $X$. It takes the shortest path between these locations. This is an (unachievable without omniscient knowledge of the object locations) upper bound on the performance of other algorithms, as it represents the fastest a scavenger hunt could possibly be completed.
	
	
	\section{Evaluation}
	\label{sec:evaluation}
	In this section we describe the simulation and real-robot experiments we conducted to evaluate the algorithms presented in Section \ref{sec:algorithms}. The algorithms' performance is evaluated in terms of travel distance and runtime.
	
	\subsection{Simulation Setup}
	
	To compare the algorithms, we implemented all seven algorithms along with a simulator which
	allows us to define a graph world, set of objects, and distribution model for the objects. Each simulated scavenger hunt randomizes the object arrangement according to the distribution model.
	\footnote{The simulator code is available at: \url{https://github.com/utexas-bwi/scavenger_hunt_api/tree/master/bwi_scavenger}}. 
	Scavenger hunts were randomly generated according to the following rules.
	
	\begin{enumerate}[itemsep=0mm,leftmargin=5mm]
		\item Hunts contain 4 objects.
		\item An object may be located at between 1 and 3 locations, with the probability of appearing at each location partitioned randomly between locations.
		\item Node locations are generated uniformly at random in an environment of size $m$ by $m$, where $m$ is 100 times the number of nodes, and the edge distances are calculated as Euclidean distances based on the node locations.
	\end{enumerate}
	The error bars in the figures indicate the 95\% confidence interval.
	\subsection{RL Simulation Results}

	In this set of experiments, we generated 10 random 8-node environments, and the DQN algorithms were trained and tested on each one separately. Note that the object locations within each environment change from trial to trial.
	
	We tested two versions of the DQN algorithm. The first, \textit{DQN}, received the distribution as part of its observation, along with the robot's location, and the list of objects that remain to be found, while the second, \textit{DQN+Map} received both the distribution and the map edge costs as part of the observation.
	
\begin{figure}[h]
		\centering
		\includegraphics[scale=0.45]{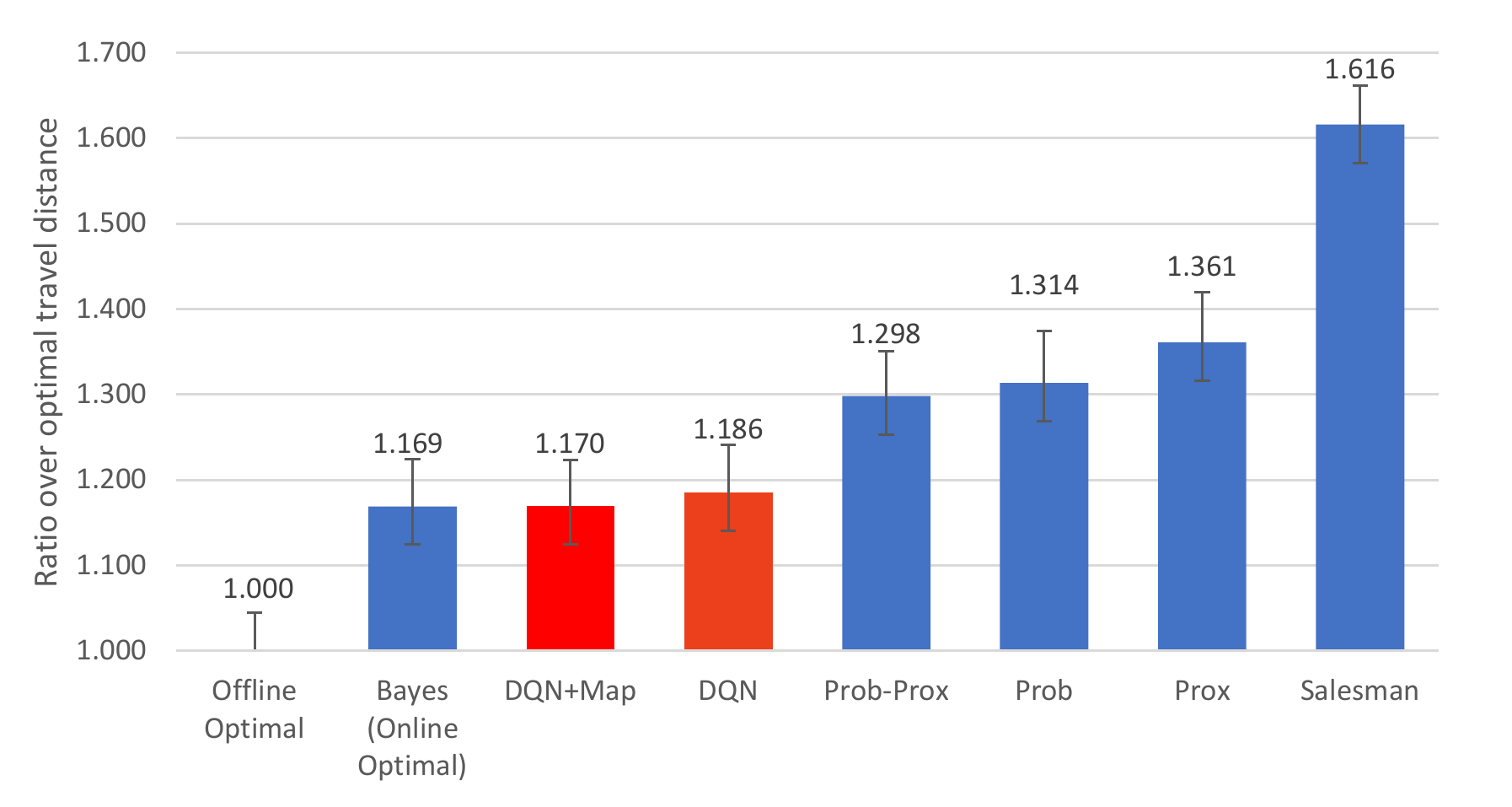}
		\caption{Comparison of algorithms on 10 environments. The DQN algorithms were trained on each environment separately and outperforms all other heuristics.}
		\label{fig:dqn}
	\end{figure}

	Figure \ref{fig:dqn} presents the ratio between the average travel distance by each algorithm to the average travel distance of the optimal solution for performing 100 hunts in each of the 10 environments. The results indicate that the DQN algorithm can outperform all other heuristics, even without knowing the edge costs. When provided with information about edge costs it can perform as well as Bayesian search with no statistical difference in performance between DQN+Map and Bayesian search with $p-value > 0.05$. Moreover, the difference between DQN and the next best heuristic, i.e. Prob-Prox, is statistically significant with $p-value = 0.006$ in a paired two sample T-test.  

	\subsection{Increasing Environment Complexity Simulation Results}
	In this section we examine the performance of the algorithms in increasingly larger environment with more nodes.
	We ran sets of trials on eight graph sizes of 3 through 10 nodes. For each graph size, each algorithm completed 100 trials of 100 scavenger hunts each, for a total of 80,000 scavenger hunts per algorithm (with the exception of exhaustive Bayesian search, which could not be run on maps of sizes 9 or 10 nodes in reasonable time due to its computational complexity. We estimate it would take approximately 8 days to run 80000 experiments on a 9 node graph). Each algorithm was given the same graph, distribution model, and object arrangement for each trial. 
	
\begin{figure}[h]
		\includegraphics[width=0.95\linewidth]{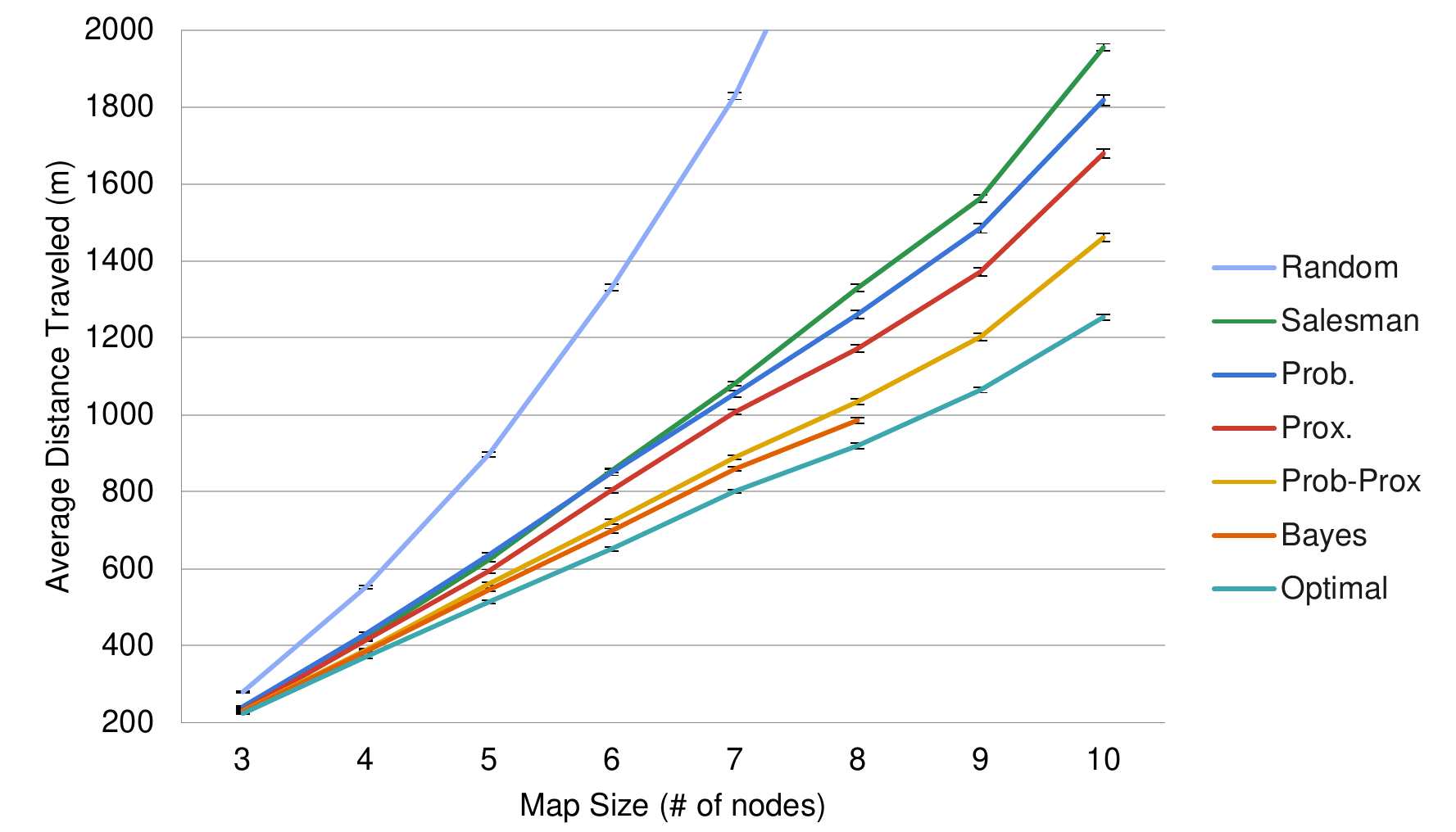}
		\caption{Simulation results comparing traveled distance, in environments with increasing complexity.}
		\label{fig:simresults}
	\end{figure}

	Figure \ref{fig:simresults} shows the performance of each algorithm in terms of its average distance traveled across various map sizes. As expected, the results are bounded by the Traveling Salesman and Offline Optimal searches. The Probability-based and Proximity-based heuristics performed almost identically, and were outperformed by the Probability-Proximity algorithm. They are all outperformed by the exhaustive Bayesian search.
	
\begin{figure}[h]
		\includegraphics[width=0.90\linewidth]{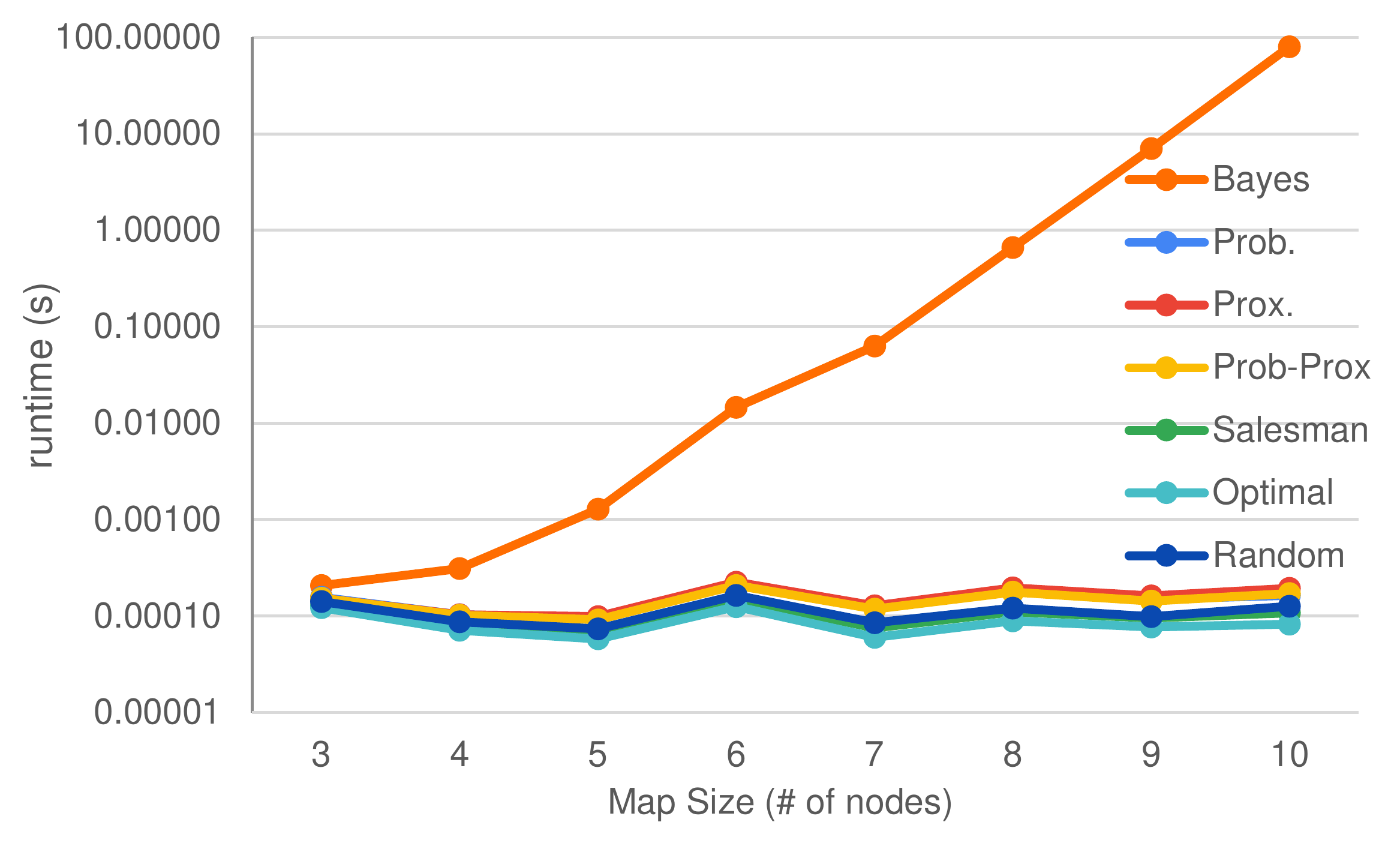}
		\caption{Run-time comparison of the algorithms on a logarithmic scale.}
		\label{fig:runtime}
	\end{figure}

	The average traveled distance for all the algorithms increases as the map grows in size. Note that the Exhaustive Bayesian search performance curve in Figure \ref{fig:simresults} is cut off at 8 nodes. For maps larger than 8 nodes, the intractable time complexity of the algorithm precludes timely completion of a full  experiment of 80,000 trials, i.e., the time to calculate the path becomes greater than 5 seconds, as seen in Figure \ref{fig:runtime}. 
	There are no statistically significant differences between the mean travel distances of Offline Optimal, Exhaustive Bayesian, and Probability-Proximity searches for environments smaller than 5 nodes, according to a two sample T-Test ($p-value>0.05$). For environments larger than 5 nodes, there is a statistically significant difference between Offline Optimal, and both Exhaustive Bayesian and Probability-Proximity searches  ($p-value< 0.05$). For environments larger that 6 nodes, there is a statistically significant difference between Exhaustive Bayesian and Probability-Proximity searches ($p-value< 0.05$).
	Figure \ref{fig:runtime} indicates that the runtime of the heuristic approaches increases linearly with the number of nodes, but the runtime of the Exhaustive Bayesian search increases exponentially (appears linear on the logarithmic scale). 

	\vspace{-10pt}

	\subsection{Robot Experiment Setup}
	
\begin{figure}[h]
		\centering
		\includegraphics[scale=0.25]{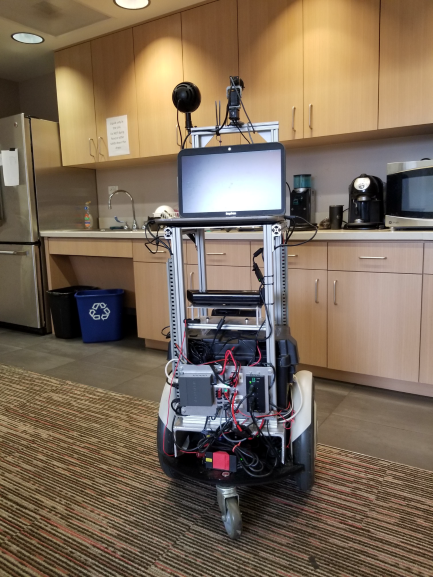}
		\caption{The robot used for the experiments.}
		\label{fig:robot}
	\end{figure}

	The Probability-Proximity, Exhaustive Bayesian, and Offline Optimal search algorithms were also tested on a real mobile robot,
	which is part of the Building-Wide Intelligence project at the University of Texas at Austin~\cite{khandelwal2017bwibots}. 
	The robot is custom built on a Segway base, with a 2D Hokuyo Lidar used for localization, and a Kinect RGBD camera for perception. A picture of the robot can be seen in Figure \ref{fig:robot}.
	
\begin{figure}[h]
		\centering
		\includegraphics[scale=0.8]{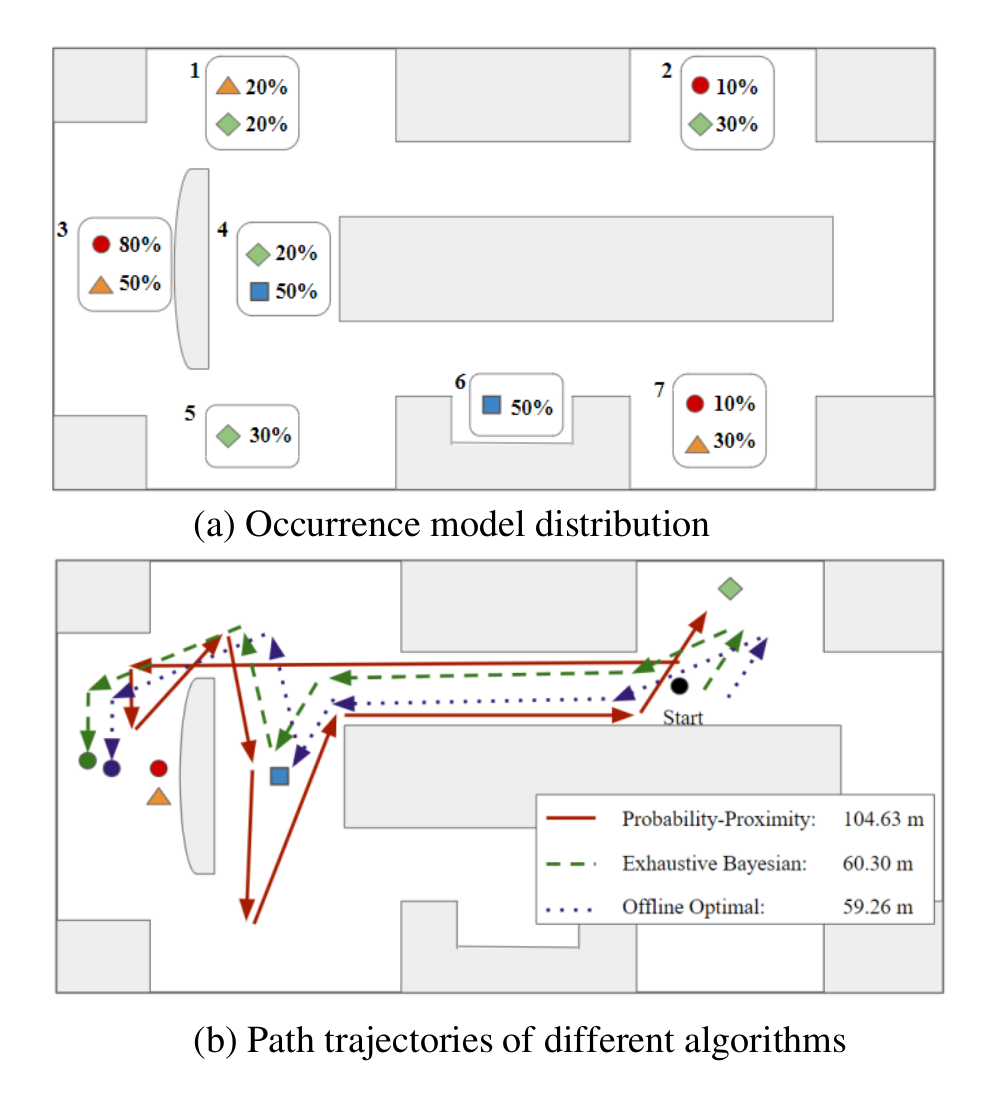}
		\caption{(a) Map of the lab space used for real-world experiments. The percentages shown are representative of the occurrence model provided to the agent. (b) Examples of the paths generated by each algorithm in a real world experiment.}
		\label{fig:paths}
	\end{figure}

	The limited computation resources on the robot resulted in long processing times by the computer vision software. To resolve that, we offboarded the object recognition process and had the robot send the feed from its Kinect V1 RGBD camera, to be processed on a separate machine. This workaround enabled the robots to scan as it was moving without having to stop and wait. 
	Scavenger hunts provided to the robot, as shown in Figure \ref{fig:paths} (a), consisted of 7 location nodes and the following distribution model:
	
	\begin{itemize}[itemsep=0mm,leftmargin=8mm]
		\item Object A occurs at node 2 10\%, node 3 80\%, and node 7 10\% of the time.
		\item Object B occurs at node 1 20\%, node 3 50\%, and node 7 30\% of the time.
		\item Object C occurs at node 1  20\%, node 2 30\%, node 4 20\% and node 5 30\% of the time.
		\item Object D occurs at node 4 50\%, and node 5 50\% of the time.
	\end{itemize}
	
	The experiment used a potted plant, teddy bear, umbrella, and soccer ball for objects A, B, C, and D, respectively. These objects were used because they were unique within the lab area and already recognized by the robot's computer vision system. It was beneficial to use uncommon items so that the robot did not accidentally find instances of the object besides those controlled by the experiment.
	In each experiment, the robot started with the same probability distribution knowledge and location as input to the algorithm. The robot determined the next best location and traveled to it. At the location, it surveyed the area and determined what objects were there. The robot updated the current probabilities to match new knowledge. It repeated these steps, until all objects were found
	
	Ten arrangements of objects were sampled according to the probabilities in the distribution model. The arrangements were replicated in the real world and the three algorithms were tested on each arrangement.
	
	Figure \ref{fig:paths} (b) shows the actual path that was generated by each algorithm for an example trial that was conducted on the real robot. 
	In this specific case, Exhaustive Bayesian produced the same path as the Offline Optimal algorithm, which resulted in approximately 60 meters in length. The Probability-Proximity path was much longer with 104.63 meters in length.

	\subsection{Robot Experiment Results}
\begin{figure}[h]
		
		\centering
		\includegraphics[width=0.60\linewidth]{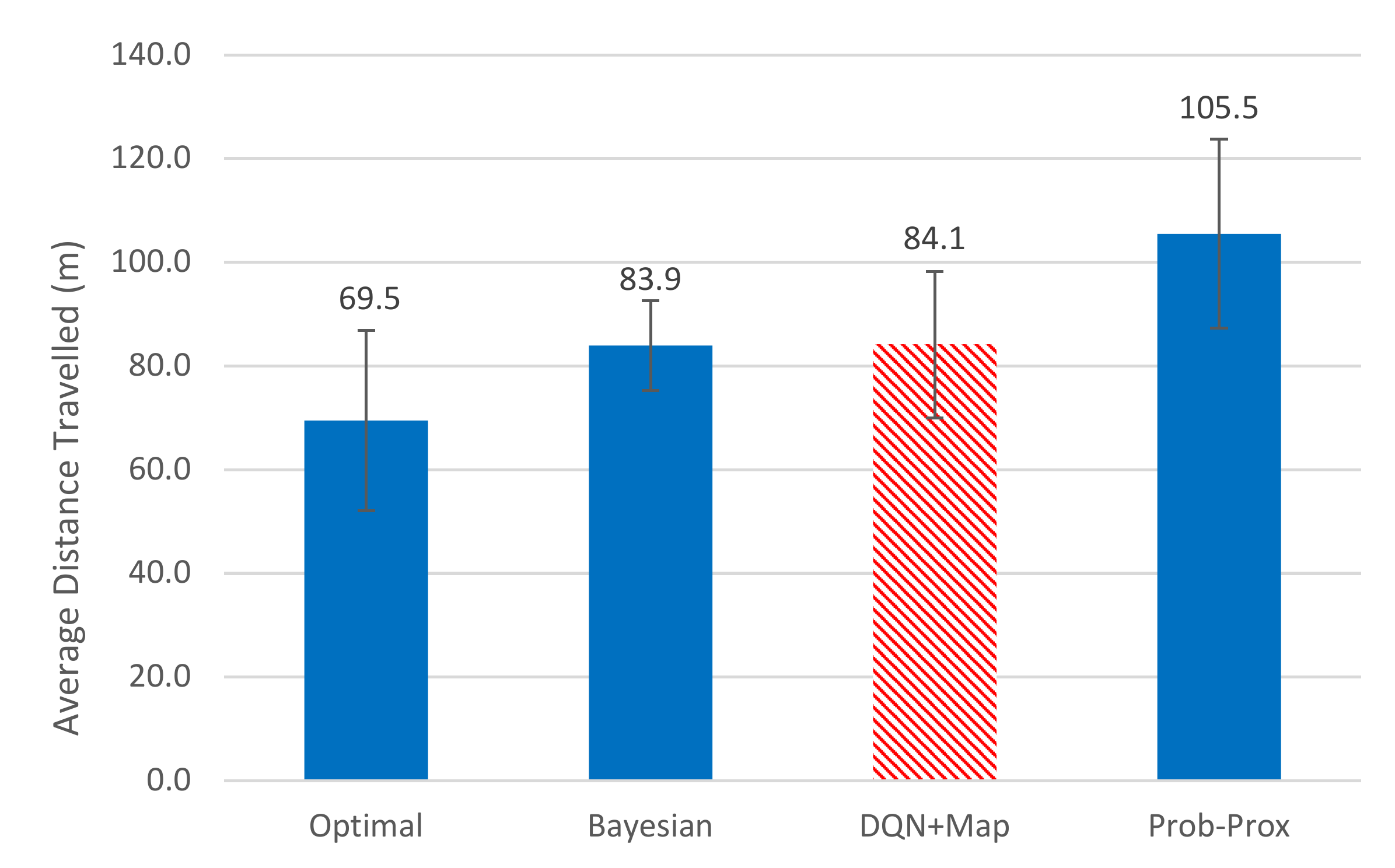}
		\caption{The average distance traveled by each algorithm in the robot experiments.}
		\label{fig:robot_distance}
	\end{figure}

	Figure \ref{fig:robot_distance} summarizes the results of the scavenger hunts that were tested on the robot. Each reported result is the average of ten trials.
	Like in simulation, Exhaustive Bayesian search outperformed the Probability-Proximity algorithm and was outperformed by Offline Optimal. Due to COVID-19 we were unable to complete the real robot experiments of the DQN algorithm. However, we ran simulated SH experiments with DQN in the simulator we created, on an environment similar to the one where the real robot experiments were run. The edge lengths in the simulated environment were exactly the lengths measured by the real robot's odometer sensor when traversing the same edges in our lab. The results indicate that DQN performs similarly to the Exhaustive Bayesian search algorithm.
	
	\section{Scavenger Hunt Website}
	\label{sec:website}
	In order to broaden the impact of this project, 
	and especially to allow researchers in other labs with other robots to participate in scavenger hunts, we also developed a website (\url{scavenger-hunt.cs.utexas.edu}) which enables users to define tasks, compose hunts, and verify proofs of completion which the robot uploads as images once it finds an object. The verification process enables the robot to maintain and update a database with object locations which it can use to plan future hunts more effectively.


	\section{Conclusion}
	\label{sec:conc}
	
	This paper defines the robot scavenger hunt problem, a stochastic traveling agent problem that involves finding objects whose appearances are probabilistic across locations. We implemented and analyzed the performance of three heuristic search algorithms for completing scavenger hunts, as well as an exhaustive Bayesian search algorithm that is optimal in expectation. All the algorithms were tested in a custom built simulated environment, while the top performing three were also tested on real world scavenger hunts. The real world hunts were conducted using a mobile robot with autonomous navigation capabilities for which we developed perception capabilities that enable it to perform scavenger hunts.
	
	Results from both simulated and real world experiments show that the Exhaustive Bayesian search algorithm outperforms all heuristic algorithms. However, it does not scale well, and is not applicable for large environments, or hunts that contain a large number of objects. The heuristic algorithms are better suited for application in large environments. 
	A key contribution of this work is in demonstrating that an RL agent trained using a standard DQN algorithm with a probability distribution model can outperform all other heuristic algorithms on the specific environment on which it was trained, and matches the performance of the exhaustive Bayesian search algorithm, without requiring much computational effort online. A natural direction for future work is to develop a learning based approach that can generalize to arbitrary environments. 
	
	Another interesting direction for future work to examine a variation of the scavenger hunt problem in which the distribution model is not known to the agent. This introduces an exploration aspect to the problem, as the agent must now explore its environment to build a distribution model while searching for objects simultaneously. This represents an exploration-exploitation tradeoff and is thus well suited for reinforcement learning.

	{\small	
		\section{Acknowledgments}
		This work has taken place in the Learning Agents Research
		Group (LARG) at UT Austin.  LARG research is supported in part by NSF
		(CPS-1739964, IIS-1724157, NRI-1925082), ONR (N00014-18-2243), FLI
		(RFP2-000), ARO (W911NF-19-2-0333), DARPA, Lockheed Martin, GM, and
		Bosch.  Peter Stone serves as the Executive Director of Sony AI
		America and receives financial compensation for this work.  The terms
		of this arrangement have been reviewed and approved by the University
		of Texas at Austin in accordance with its policy on objectivity in
		research.
	}
	
	\bibliographystyle{abbrv}
	\bibliography{references}  
	
\end{document}


\maketitle
	
	\appendix

	
	\section{Website}
	\label{sec:website_appendix}
	In order to facilitate the definition of scavenger hunts, the verification of \textit{proofs of completion} ("proofs" is short), and especially to allow researchers in other labs with other robots to participate in scavenger hunts, we also developed a website which enables users to define tasks, compose hunts and verify proofs. The verification process enables the robot to maintain and update a database with object locations which it can use to plan future hunts more effectively.
	The framework we developed is publicly available and our goal is to have other research labs join and run scavenger hunts on their robots using this framework.

	
	\begin{figure*}[hbt]
		\centering
		\includegraphics[scale=0.89]{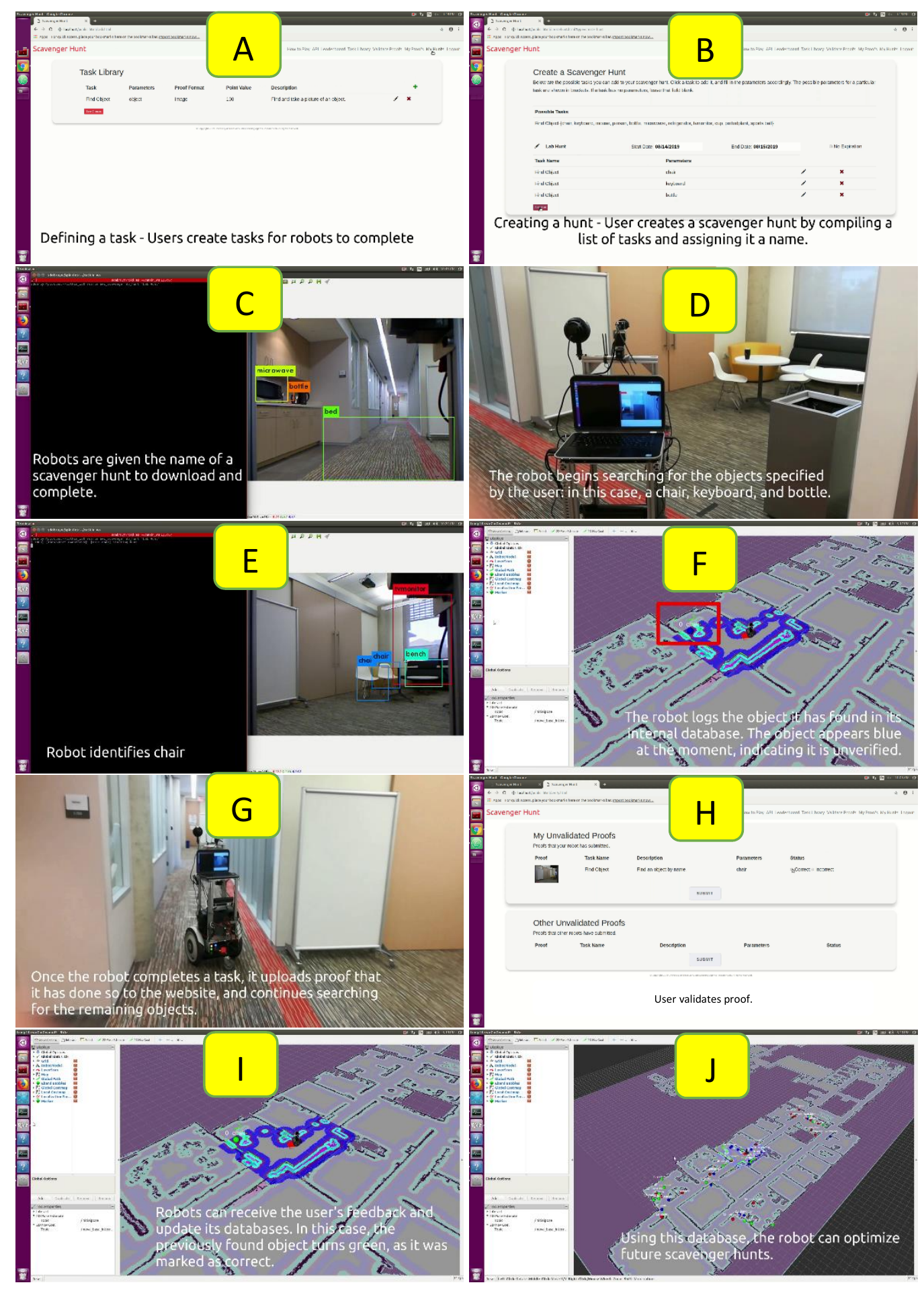}
		\vspace{0pt}
		\caption{This figure displays the workflow of a scavenger hunt.}
		\label{fig:website}
	\end{figure*}
	
	The workflow of a scavenger hunt as depicted in figure \ref{fig:website} contains 10 stages:
	\begin{enumerate}[label=(\Alph*)]
		\item User can define tasks such as "Find an object and take a picture of it".
		\item The user can then compose scavenger hunts containing a set of tasks.
		\item The robot downloads the hunt.
		\item The robot starts searching for the objects in the hunt (chair , keyboard, and bottle).
		\item The robot identifies a chair.
		\item The robot logs the location of the object in its database - currently unverified (colored blue). 
		\item The robot uploads proof of finding the object to the website, in this case a picture of the item it had found, and continues searching for the rest of the objects.
		\item The user verifies that the proof is correct. 
		\item The robot updates its database by changing the color of the object location to green to indicate that it is verified.
		\item The robot maintains the database of previous object locations and uses it to plan its next hunts more efficiently.
	\end{enumerate}
	
	\newpage
	\section{Algorithm Pseudocode }
	\label{sec:algotithms_pseudocode}
	The pseudocode for all the solution algorithms described in Section 4 are presented in this section to fully specify our algorithms and to aid replicability. 
	\begin{algorithm}[h]
		\caption{Proximity-Based Search}
		\label{alg:prox}
		\begin{algorithmic}[1]
			\REQUIRE{A graph $G = (N,E)$, a set of objects $O$, a distribution $D$, start location node $n_0$}
			\STATE{Initialize $n_0$, $Y_0$ and $\tau_0$ with $t = 0$}
			\WHILE{$Y_{t, i} \neq 0, \forall i \in [k]$}
			\STATE{$n_{t+1}$ $\xleftarrow{}$ $\argmin\limits_{n \in N} \{e_{n_t, n}|\exists~o \in O, {p_t}_o^{n} > 0\}$}
			\STATE{Travel to $n_{t+1}$}
			\STATE{Update $D$, $Y_{t+1}$ and $\tau_{t+1}$ based on the occurrence of objects at $n_{t+1}$}
			\STATE{$t = t+1$}
			\ENDWHILE  
		\end{algorithmic}
	\end{algorithm}
	
	\begin{algorithm} [h]
		
		\caption{Probability-Based Search}
		\label{alg:prob}
		\begin{algorithmic}[1]
			\REQUIRE{A graph $G = (N,E)$, a set of objects $O$, a distribution $D$, start location node $n_0$}
			\STATE{Initialize $n_0$, $Y_0$ and $\tau_0$ with $t = 0$}
			\WHILE{$Y_{t, i} \neq 0, \forall i \in [k]$}
			\STATE{$n_{t+1}$ $\xleftarrow{}$ $\argmax\limits_{n\in N} \{1-\prod\limits_{o \in O} (1-p_o^n)\}  $}
			\STATE{Travel to $n_{t+1}$}
			\STATE{Update $D$, $Y_{t+1}$ and $\tau_{t+1}$ based on the occurrence of objects at $n_{t+1}$}
			\STATE{$t = t+1$}
			\ENDWHILE 
		\end{algorithmic}
		
	\end{algorithm}
	
	\begin{algorithm} [h]
		
		\caption{Probability-Proximity Search}
		\label{alg:prob_prox}
		\begin{algorithmic}[1]
			\REQUIRE{A graph $G = (N,E)$, a set of objects $O$, a distribution $D$, start location node $n_0$}
			\STATE{Initialize $n_0$, $Y_0$ and $\tau_0$ with $t = 0$}
			\WHILE{$Y_{t, i} \neq 0, \forall i \in [k]$}
			\STATE{$n_{t+1}$ $\xleftarrow{}$ $\argmax\limits_{n\in N} \frac{1-\prod\limits_{o \in O} (1-p_o^{n})}{e_{n_t, n}}$}
			\STATE{Travel to $n_{t+1}$}
			\STATE{Update $D$, $Y_{t+1}$ and $\tau_{t+1}$ based on the occurrence of objects at $n_{t+1}$}
			\STATE{$t = t+1$}
			\ENDWHILE  
		\end{algorithmic}
		
	\end{algorithm}	
	
	\begin{algorithm} [h]
		
		\caption{DQN Algorithm}
		\label{alg:dqn}
		\begin{algorithmic}[1]
			\REQUIRE{A graph $G = (N,E)$, a set of objects $O$, a distribution $D$, start location node $n_0$, a trained policy $\pi(a|s)$}
			\STATE{Initialize $n_0$, $Y_0$ and $\tau_0$ with $t = 0$}
			\WHILE{$Y_{t, i} \neq 0, \forall i \in [k]$}
			\STATE{$s$ $\xleftarrow{}$ build\_observation(G,D)}
			\STATE{$n_{t+1}$ $\xleftarrow{}$ $\argmax\limits_{a \in N} \pi(a|s)$}
			\STATE{Travel to $n_{t+1}$}
			\STATE{Update $D$, $Y_{t+1}$ and $\tau_{t+1}$ based on the occurrence of objects at $n_{t+1}$}
			\STATE{$t = t+1$}
			\ENDWHILE  
		\end{algorithmic}
		
	\end{algorithm}	
	
	\begin{algorithm} [h]
		
		\caption{Exhaustive Bayesian Search}
		\label{alg:bayesian}
		\begin{algorithmic}[1]
			\REQUIRE{A graph $G = (N,E)$, a set of objects $O$, a distribution $D$, start location node $n_0$}
			\STATE{Initialize $n_0$, $Y_0$ and $\tau_0$ with $t = 0$}
			\WHILE{$Y_{t, i} \neq 0, \forall i \in [k]$}
			\STATE{best\_path=$\emptyset$}
			\STATE{best\_cost=$\emptyset$}
			\FOR{path in all the possible paths}
			
			\STATE{expected\_cost = $\sum\limits_{X \in N^m} D(X|\tau_t) *$ compute\_cost(path, $X$)}
			
			\IF{expected\_cost $<$ best\_cost}
			\STATE{best\_path=path}
			\STATE{best\_cost=expected\_cost}
			\ENDIF
			\ENDFOR
			\STATE{$n_{t+1}$ $\xleftarrow{}$ first node in best\_path}
			\STATE{Travel to node $n_{t+1}$}
			\STATE{Update $D$, $Y_{t+1}$ and $\tau_{t+1}$ based on the occurrence of objects at $n_{t+1}$}
			\STATE{$t = t+1$}
			\ENDWHILE
			
		\end{algorithmic}
		
	\end{algorithm}	
	
	\begin{algorithm} [h]
		
		\caption{Salesman Search}
		\label{alg:salesman}
		\begin{algorithmic}[1]
			\REQUIRE{A graph $G = (N,E)$, a set of objects $O$, a distribution $D$, start location node $n_0$}
			\STATE{Initialize $n_0$, $Y_0$ and $\tau_0$ with $t = 0$}
			\STATE{shortest\_path $\xleftarrow{}$ compute\_shortest\_path($N$, $E$)\hfill\COMMENT{shortest path that visits all the nodes in $N$}
			}
			\WHILE{$Y_{t, i} \neq 0, \forall i \in [k]$}
			\STATE{$n_{t+1}$ $\xleftarrow{}$ next\_node\_in\_path(shortest\_path, $n_t$)}
			\STATE{Travel to $n_{t+1}$}
			\STATE{Update $D$, $Y_{t+1}$ and $\tau_{t+1}$ based on the occurrence of objects at $n_{t+1}$}
			\STATE{$t = t+1$}
			\ENDWHILE  
		\end{algorithmic}
		
	\end{algorithm}	
	
	\begin{algorithm} [h]
		
		\caption{Offline Optimal Search}
		\label{alg:optimal}
		\begin{algorithmic}[1]
			\REQUIRE{A graph $G = (N,E)$, a set of objects $O$, objects locations $X$, a distribution $D$, start location node $n_0$}
			\STATE{Initialize $n_0$, $Y_0$ and $\tau_0$ with $t = 0$}
			\STATE{shortest\_path $\xleftarrow{}$ compute\_shortest\_path($X$, $E$)\hfill\COMMENT{shortest path that visits all the nodes in $X$}
			}
			\WHILE{$Y_{t, i} \neq 0, \forall i \in [k]$}
			\STATE{$n_{t+1}$ $\xleftarrow{}$ next\_node\_in\_path(shortest\_path, $n_t$)}
			\STATE{Travel to $n_{t+1}$}
			\STATE{Update $D$, $Y_{t+1}$ and $\tau_{t+1}$ based on the occurrence of objects at $n_{t+1}$}
			\STATE{$t = t+1$}
			\ENDWHILE  
		\end{algorithmic}
		
	\end{algorithm}	
	\newpage
	
	\newpage
	\section{Learning Curve of DQN}
	\begin{figure}[h]
		\centering
		\includegraphics[scale=0.7]{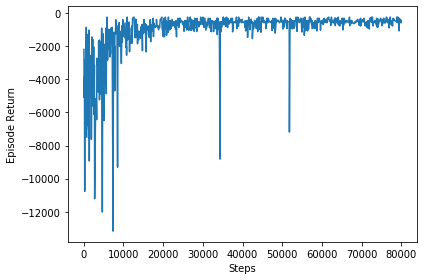}
		\caption{Learning curve of DQN on one of the environment}
		\label{fig:ep_return}
	\end{figure}

	\section{Robot Experiments Additional Data}
	\label{sec:robot_experiment_dqn}
\begin{figure}[h]
		\centering
		\includegraphics[scale=0.7]{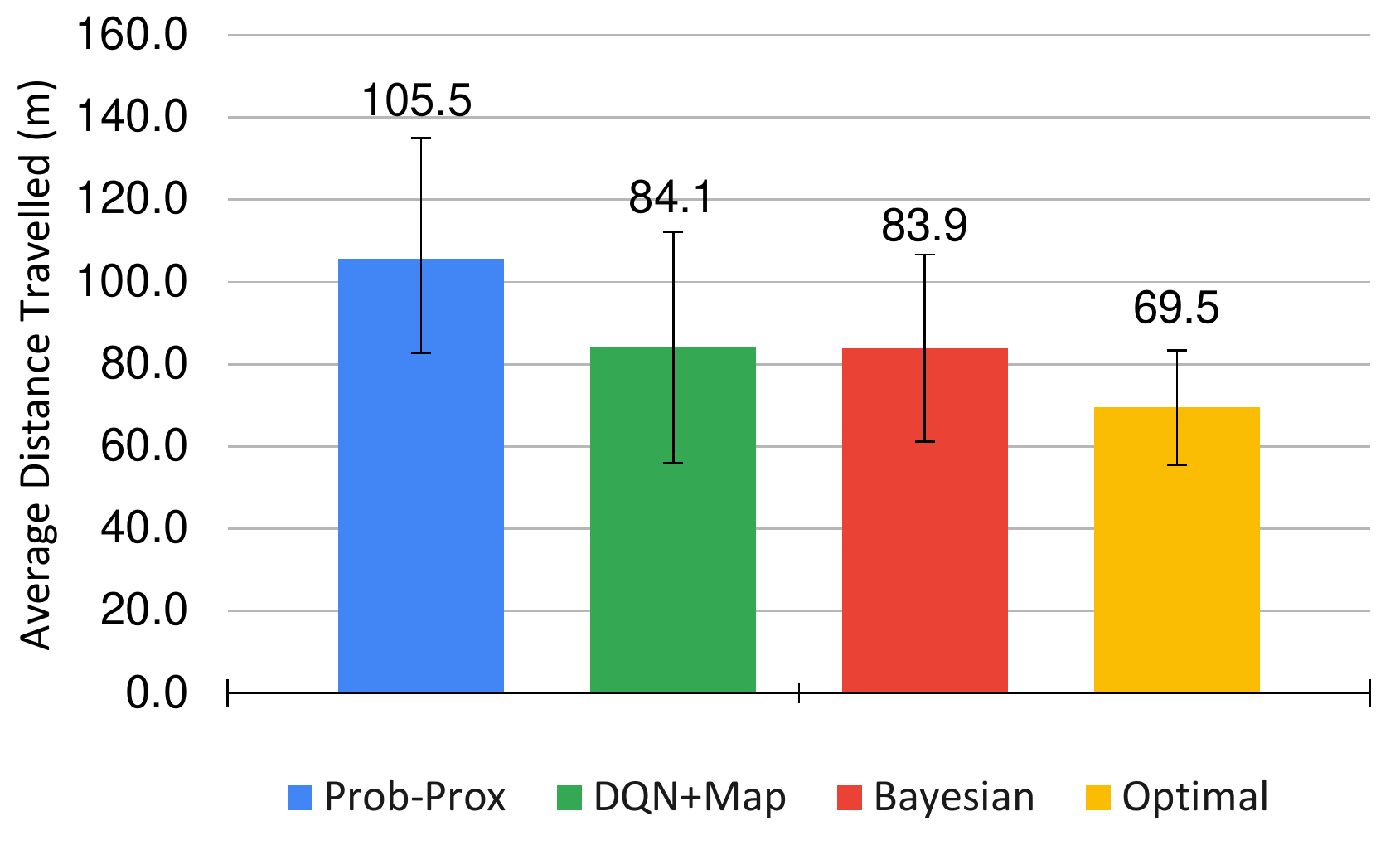}
		\caption{Estimated performance of DQN on real-world experiments. }
		\label{fig:dqn_robot}
	\end{figure}

	Due to the COVID-19 lock-down in our university, we were unable to complete all the real robot experiments we intended to perform. Specifically, the DQN algorithm's performance was not evaluated on a real robot. However, we ran simulation experiments on the same environment as the other robot evaluations for DQN.
	In this supplementary material we present this estimated result of the RL agent, as we expect it to be on a real robot. 
	Figure \ref{fig:dqn_robot} presents the DQN+map algorithms performance on a simulated robot as compared to the other three algorithms which were tested on the robot. The results indicate that the DQN agent performs similarly to the Exhaustive Bayesian search algorithm. The results of the Optimal, Bayesian ans Prob-Prox algorithms were generated from the real robot's paths as presented in Figure 4(b) in the main paper.
	
	\section{Acquiring the object distribution}
	\label{sec:acquiring_distribution}
	A preliminary stage to solving a scavenger hunt problem is acquiring the object distribution
	The acquisition process is done in the following way. The robot starts with a uniform distribution for all the objects over all the nodes. With every visit to a node, the probability is updated based on the  observation in that location: found objects will have 0 probability at all other locations, and objects that were not found will have their probability for all other possible locations redistributed to sum to 1. At the end of the hunt, the robot uploads the proofs to the website. Once those proofs are validated, the robot updates its knowledge base, incrementing the number of observations of found objects at the locations where they were found. 
	In this paper we only focus on the evaluating the algorithms' ability to plan given a distribution, and leave the acquisition process for future work. 

	\section{Notes on the simulation environment}
	\label{sec:simulation_env}	
	The reason that the experiments in the simulator had fully connected graphs is as follows.
	When the robot chooses a node to visit next, the assumption is that it does not take observations from the nodes along the path to the chosen node. Under this assumption, a path between non-neighboring nodes in a partially connected graph can be represented as a pseudo edge with length equal to the minimum distance path between the two nodes.  Hence, under this assumption, any partially connected graph can be represented as a fully connected graph, and that is the reason we only simulated fully connected graphs.